\pdfoutput=1
\documentclass[11pt]{article}

\usepackage{acl}

\usepackage{times}
\usepackage{latexsym}
\usepackage{multirow}
\usepackage{graphicx}
\usepackage[T1]{fontenc}
\usepackage{amsmath}
\usepackage{babel}
\usepackage{booktabs}
\usepackage{array}
\usepackage{caption}
\usepackage{subcaption}
\usepackage{rotating}

\usepackage[utf8]{inputenc}

\newcommand{\method}{{\fontfamily{SourceCodePro-TLF}\selectfont SynthDST}}

\newcommand{\refine}[1]{\textcolor{black}{#1}}

\newcommand{\hint}[1]{\footnote{\hspace{-1.8em}% from \@makefntext
  \rotatebox[origin=c]{180}{\parbox[t]{\linewidth}%
    {\advance\parfillskip by 1.8em\relax Hint: #1}}%
  \vskip\ht\strutbox}}

\title{\textit{\method}: Synthetic Data is All You Need for Few-Shot Dialog State Tracking}

\author{
    Atharva Kulkarni$^{1}$\thanks{\rule{1ex}{0ex}Work done during internship at Apple Inc.}$\;$,  Bo-Hsiang Tseng$^{2}$, Joel Ruben Antony Moniz$^{2}$, \\
    \textbf{Dhivya Piraviperumal$^{2}$}, \textbf{Hong Yu$^{2}$}, \textbf{Shruti Bhargava$^{2}$} \\
    $^1$Language Technologies Institute, Carnegie Mellon University, \\
    $^2$Apple Inc.\\
    \texttt{atharvak@cs.cmu.edu}, \\ \texttt{\{bohsiang\_tseng, joelrubenantony\_moniz,} \\
    \texttt{\{dhivyaprp, hong\_yu, shruti\_bhargava\}@apple.com}
}

\begin{document}
\maketitle

\begin{abstract}

% The advent of Large Language Models (LLMs) has empowered ML practitioners to use their in-context learning capabilities to improve dialogue systems' Dialog State Tracking (DST) performance.
In-context learning with Large Language Models (LLMs) has emerged as a promising avenue of research in Dialog State Tracking (DST). However, the best-performing in-context learning methods involve retrieving and adding similar examples to the prompt, requiring access to labeled training data. Procuring such training data for a wide range of domains and applications is time-consuming, expensive, and, at times, infeasible. While zero-shot learning requires no training data, it significantly lags behind the few-shot setup. Thus, `\textit{Can we efficiently generate synthetic data for any dialogue schema to enable few-shot prompting?}' Addressing this question, we propose \method, a data generation framework tailored for DST, utilizing LLMs. Our approach only requires the dialogue schema and a few hand-crafted dialogue templates to synthesize natural, coherent, and free-flowing dialogues with DST annotations. Few-shot learning using data from {\method} results in $4-5\%$ improvement in Joint Goal Accuracy over the zero-shot baseline on MultiWOZ 2.1 and 2.4. Remarkably, our few-shot learning approach recovers nearly $98\%$ of the performance compared to the few-shot setup using human-annotated training data\footnote{Our synthetic data and code can be accessed at \\ \url{ https://github.com/apple/ml-synthdst}.}.
    % Zero-shot evaluation for dialogue state tracking with LLMs has been reported to have lower performance compared to the few-shot setups. Prior works have shown that augmenting the prompt with examples from training data that are similar to the test instance boosts performance significantly. However, collection of training data can be time consuming and expensive. In this work, we aim to address this issue. We explore the synthesis of such data with our proposed framework {\method}. We particularly focus on two turn dialogues with diverse state changes. {\method} can be extended to any new domain by providing just the domain schema. Using the synthetic examples generated by {\method} in the prompt for LLMs results in an 8\% absolute improvement in Joint Goal Accuracy compared to the zero-shot setting on MultiWOZ 2.1. Moreover, despite not relying on any human annotations, the performance comes close to the few shot evaluation using human annotated training data.
\end{abstract}

\section{Introduction}

% \vspace{-3mm}
% What is dialogue state tracking and what are the current bottlenecks. 1) Not access to large amount of data. 2) Getting data is difficult. 3) Annotations errors. 4) Do not do well on different distrubtions
% \blfootnote{$^1$Work done during internship at Apple.} 
\textit{Dialogue State Tracking (DST)} is an integral task in task-oriented dialogue systems that predicts the user intentions for each turn by mapping them to predefined slot-value pairs \cite{henderson2015machine}. DST systems capture important information essential to model the downstream dialogue policy and help generate actionable responses \cite{jacqmin-etal-2022-follow}. Prior literature has typically framed DST as either a multi-class classification task \cite{henderson-etal-2014-word,mrksic-etal-2017-neural, wu-etal-2020-tod, Chen_Lv_Wang_Zhu_Tan_Yu_2020} or a sequence-to-sequence learning task \cite{wu-etal-2019-transferable, kim-etal-2020-efficient, NEURIPS2020_e9462095, lee-etal-2021-dialogue, shin-etal-2022-dialogue}. With the rise of Large Language Models (LLMs), various techniques have been proposed to harness their emergent capabilities for dialogue state tracking \cite{hu-etal-2022-context, chen-etal-2023-stabilized, heck-etal-2023-chatgpt, king-flanigan-2023-diverse, yang2023dual}. 

\begin{figure}[t!]
    \centering
    \includegraphics[width=\columnwidth, height=4.5cm]{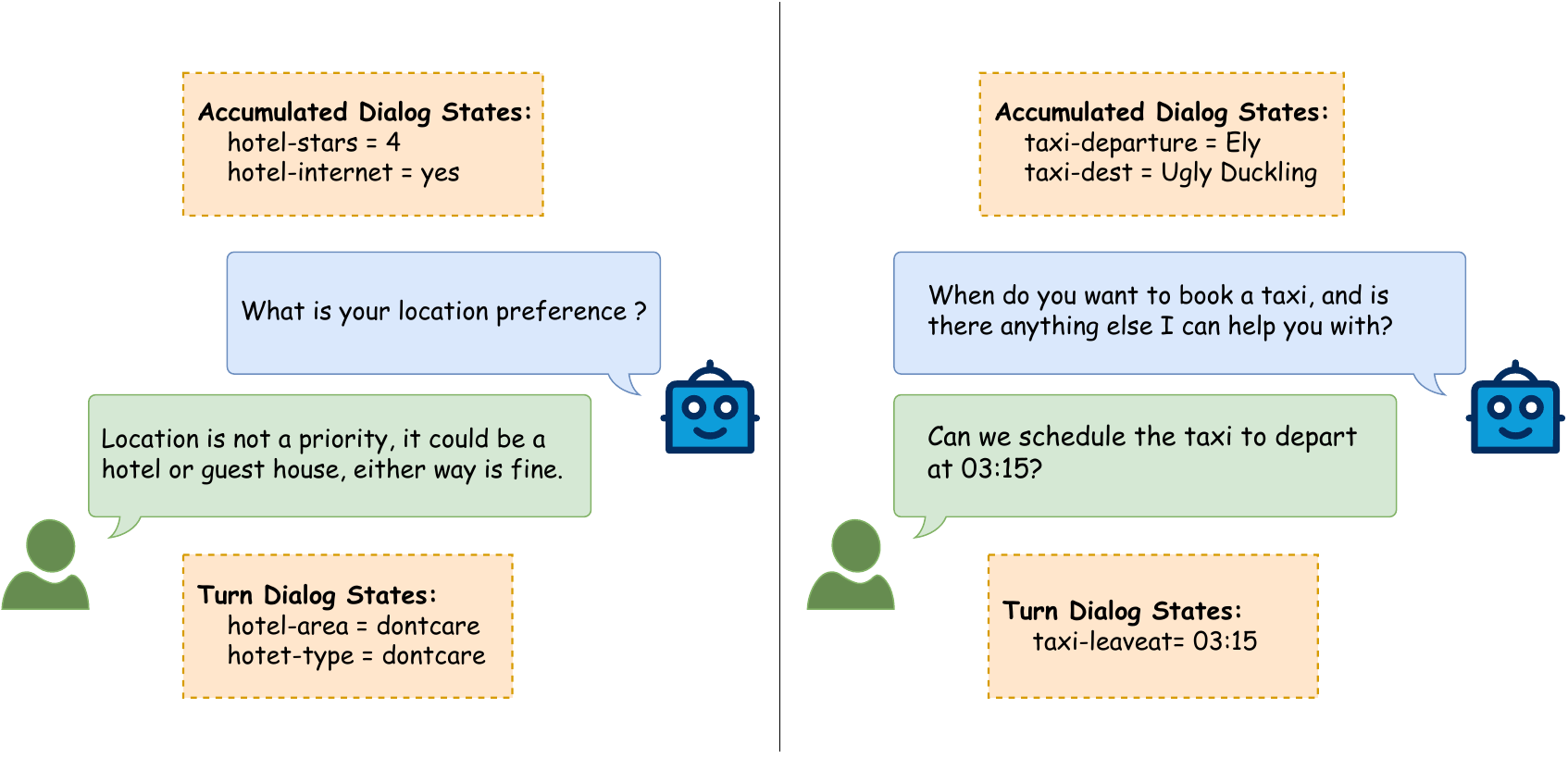}
    \caption[check]{One of these is a dialog generated by \method. Each dialog contains conversation history (as accumulated dialog states), system turn, user turn, and current turn's dialog states. Can you guess which dialog is synthetically generated by \method?\footnotemark.}
    \vspace{-5mm}
    \label{fig:example}
\end{figure}
\footnotetext{The right example is synthetically generated.}
% \footnotetext{Answer:\hspace{-6em}% from \@makefntext
%   \rotatebox[origin=c]{180}{\parbox[t]{\linewidth}%
%     {\advance\parfillskip by 1.8em\relax The right example is synthetically generated.}}%
  % \vskip\ht\strutbox}
  
% \textit{Dialogue State Tracking (DST)} is an integral component in task-oriented dialogue systems that extracts user intentions for each turn by mapping them to predefined slots values \cite{henderson2015machine}. DST captures important information that is essential to model the downstream dialogue policy and to generate actionable responses \cite{jacqmin-etal-2022-follow}. Prior literature in dialogue systems has typically framed DST as either a multi-class classification problem \cite{henderson-etal-2014-word,mrksic-etal-2017-neural, wu-etal-2020-tod, Chen_Lv_Wang_Zhu_Tan_Yu_2020} or as a generative sequence-to-sequence learning task \cite{wu-etal-2019-transferable, kim-etal-2020-efficient, NEURIPS2020_e9462095, lee-etal-2021-dialogue, shin-etal-2022-dialogue}. However, with the rise of Large Language Models (LLMs), various techniques have been proposed to harness their emergent capabilities for dialogue state tracking \cite{hu-etal-2022-context, chen-etal-2023-stabilized, heck-etal-2023-chatgpt, king-flanigan-2023-diverse, yang2023dual}. 

Most approaches for DST necessitate access to gold-standard human-annotated data for supervised fine-tuning \cite{wu-etal-2019-transferable, shin-etal-2022-dialogue} or retrieval-based in-context learning \cite{hu-etal-2022-context, king-flanigan-2023-diverse}. This comes with four main drawbacks. First, curating fine-grained utterance-level annotated dialogue data in a \textit{Wizard-of-Oz} / \textit{human-to-human conversation} setup (e.g., MultiWOZ) is both time-consuming and expensive \cite{budzianowski-etal-2018-multiwoz}. 
% Alternatively, \textit{simulation-based} or \textit{machine-to-machine conversation} frameworks (e.g.,  M2M) present a less expensive data generation approach; however, they provide limited complexity and linguistic diversity \cite{shah2018building, Rastogi_Zang_Sunkara_Gupta_Khaitan_2020}. 
Second, many DST benchmarks contain incorrect annotations \cite{ye-etal-2022-multiwoz}, which can hinder learning and may introduce spurious biases \cite{qian-etal-2021-annotation}. Third, nearly all DST datasets are confined to a limited number of domains. Training on these datasets limits the models' ability to generalize to unseen domains, thereby hampering their suitability for real-world deployment \cite{dingliwal-etal-2021-shot}. Fourth, real-world applications may need to regularly add new domains or modify existing schemas. However, iterating on data collections may pose a significant challenge \cite{jacqmin-etal-2022-follow}. 
% A model trained on static datasets becomes ineffective as the expected data distribution evolves over time. 

% Most approaches for DST necessitate access to gold annotated training data for supervised fine-tuning or retrieval-based in-context learning \cite{hu-etal-2022-context}. This comes with three main drawbacks. First, collecting fine-grained annotated dialogue data in a \textit{Wizard-of-Oz}, or \textit{human-to-human conversation} setup (e.g., MultiWOZ), is time consuming and monetarily expensive \cite{budzianowski-etal-2018-multiwoz}. One can instead opt for \textit{simulation-based} or \textit{machine-to-machine conversation} data generation approaches (e.g.,  M2M) \cite{shah2018building, Rastogi_Zang_Sunkara_Gupta_Khaitan_2020}, however the current techniques provide limited complexity and diversity. Second, many DST benchmarks contain noisy annotations \cite{ye-etal-2022-multiwoz}, which can hinder learning and may introduce spurious biases \cite{qian-etal-2021-annotation}. Third, nearly all DST datasets are confined to limited domains, and training on them offers limited generalizability. Real world applications may need to support new domains or modify domain schemas frequently. However, collecting and updating data regularly can be a significant bottleneck \cite{dingliwal-etal-2021-shot, jacqmin-etal-2022-follow}.

While zero-shot prompting of LLMs using only the dialogue schema provides a data-less approach for DST, it under-performs compared to the retrieval-based few-shot prompting that adds semantically similar training examples in the prompt \cite{hu-etal-2022-context, king-flanigan-2023-diverse}. Given these challenges, one may wonder: \textit{`How can we leverage LLMs' in-context learning capabilities when we do not have access to annotated training data?'} Or conversely, \textit{`Can we efficiently generate synthetic data for any dialogue schema to enable few shot prompting?'} In this work, we aim to answer this.

% While zero-shot prompting using only the dialogue schema provides a data-less approach for DST, it under-performs compared to the retrieval-based few-shot prompting that adds semantically similar training examples in the prompt \cite{hu-etal-2022-context}. Given these challenges, one may ask: \textit{`Can we use retrieval-based few shot prompting even in the absence of annotated training data'} or in turn \textit{`Can we efficiently generate synthetic data for any dialogue schema to enable few shot prompting?'}. In this work, we aim to answer this.

\begin{figure*}[t!]
    \centering
    \includegraphics[width=\textwidth]{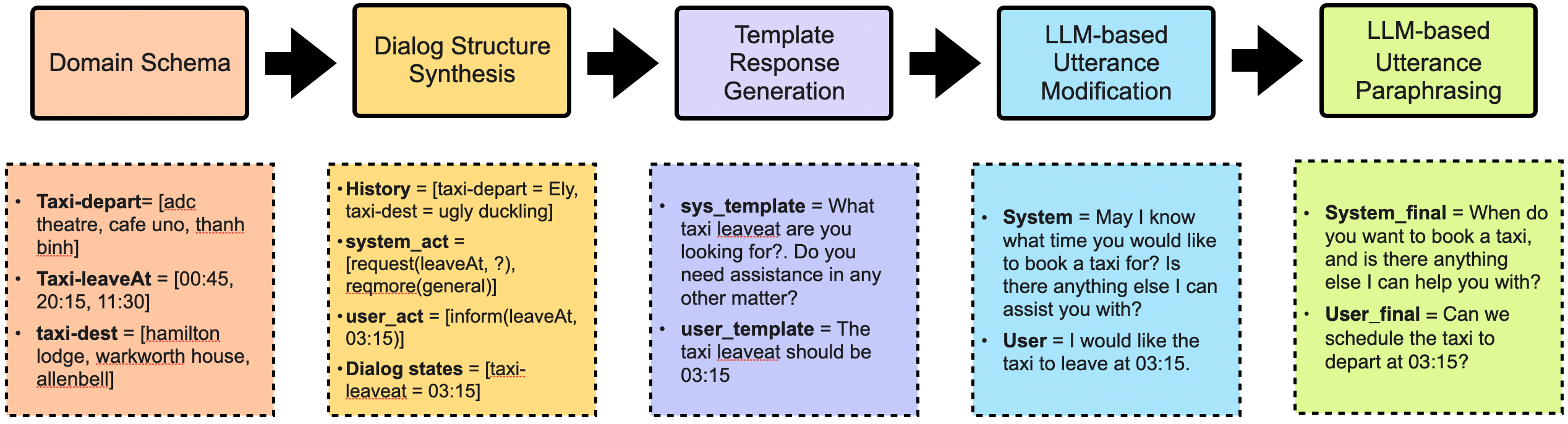}
    \caption{Overall pipeline of \method\ for synthetic dialog generation}
    \label{fig:pipeline}
\end{figure*}

We introduce \method, an LLM-based approach for generating dialogues with dialog state annotations. \method\ takes a dialogue schema as input and outputs four objects: the conversation state, the next system response, the next user response, and the updated conversation state. For this, it uses predefined intents and intent transitions (Table \ref{tab:abstract_model}), along with hand-crafted templates (Tables \ref{tab:templates_system}, \ref{tab:templates_user}). The pipeline is detailed in Figure \ref{fig:pipeline}, and an example can be seen in Figure \ref{fig:example}. It first programmatically generates raw data for the four output objects. Then, it transforms the raw intents into sentences with templates and further paraphrases them into natural language using LLMs. We evaluate \method\ using the IC-DST framework \cite{hu-etal-2022-context} on MultiWOZ 2.1 \cite{eric-etal-2020-multiwoz} and 2.4 \cite{ye-etal-2022-multiwoz}. Our results show a $4-5\%$ improvement over the zero-shot baseline on both datasets. Moreover, few-shot learning with \method\ data achieves approximately $98\%$ and $95\%$ of the performance when using training data for MultiWOZ 2.1 and 2.4, respectively. In summary, our contributions are two-fold:
\begin{itemize}
    \item We propose \method, a scalable domain agnostic framework for generating synthetic dialogue data with dialog state annotations.
    \item We empirically demonstrate that retrieval-based few-shot prompting with \method's synthetic data surpasses both the zero-shot and random few-shot learning baselines. Moreover, it reaches close to the few-shot prompting performance with human-annotated training data.
    % \item Our approach with synthetic data helps unveil and remove annotation biases that are present in the training datasets.
\end{itemize}

\section{Related Work}
% \atharva{TODO}

\subsection{Synthetic Data Generation for Dialog}
The advent of large language models has brought about a significant transformation in synthetic data generation.
% \citet{xiang-etal-2022-asdot} and \citet{ye-etal-2022-progen} demonstrate that LLMs can effectively generate synthetic data for any arbitrary domain and task. Consequently, numerous attempts have been made to generate synthetic dialogue corpora using LLMs. 
\textit{LAD} \cite{mehri-etal-2022-lad} generates linguistically diverse synthetic dialogues by imposing structural constraints on prompts for intent prediction, slot filling, and next action prediction. \textit{RSODD} \cite{bae-etal-2022-building} adopts a human-in-the-loop approach to craft role-specific open-domain dialogues. Specifically, it takes role specification and examples designed by dialogue developers to generate artificial conversations, followed by human editing. On similar lines, \citet{chen-etal-2023-places} introduced \textit{PLACES}, a framework utilizing topic information, background context, and expert-written conversations as in-context examples for synthetic dialogue generation. \textit{Synergy} \cite{peng2021synergy} adopts a different approach by modifying simulated dialogue sketches, each comprising multi-turn dialogue actions and belief states. A natural language generation module transforms these actions into natural language. Lastly, \textit{DIALOGIC} \cite{li-etal-2022-controllable} presents a controllable dialogue simulation method that generates DST-annotated dialogues using a seed corpus.

% \shruti{specify how ours is similar and different from these}
In summary, the NLP community has shown a growing interest in synthetic data generation for dialogue applications. However, frameworks like \textit{RSODD}, \textit{PLACES}, and \textit{Synergy} demand a level of human supervision and lack slot-value annotations, rendering them unsuitable for DST. While \textit{DIALOGIC} generates synthetic data with dialog state annotations, it has limited coverage of dialogue acts, needs human intervention for annotation correction, and necessitates a seed corpus. 
% Building upon prior research that synthesized dialogue data using LLMs \cite{li-etal-2022-controllable, mehri-etal-2022-lad, chen-etal-2023-places, deng2023prompting, chen2023controllable}, 
Addressing these challenges, \method\ provides more control over the generated dialogues by grounding them in dialogue states. Moreover, \method\ does not require human intervention in filtering or editing the synthetic data, facilitating greater scalability.

\subsection{Zero and Few-Shot Learning for DST}
Significant research is dedicated to dialogue state tracking with in-context learning. \citet{lin-etal-2021-leveraging} proposed a zero-shot cross domain DST method by prompting the T5 model \cite{JMLR:v21:20-074} with slot descriptions. \citet{madotto2021few} assessed different language models for DST through prompt-based few-shot learning. Other approaches, such as UnifiedSKG \cite{xie-etal-2022-unifiedskg} and InstructDial \cite{gupta-etal-2022-instructdial}, introduce multi-tasked and instruction-tuned variants of T5 and BART, which exhibit strong zero-shot DST performance. IC-DST \cite{hu-etal-2022-context} frames DST as a text-to-SQL problem using the codex version of GPT-3. RefPyDST framework \cite{king-flanigan-2023-diverse} formulates DST as a Python programming task, retrieves more diverse in-context examples, and introduces a novel re-weighting method during decoding. \citet{heck-etal-2023-chatgpt} provide empirical evidence that ChatGPT can yield competitive results in DST without any complex prompting. More recently, a dual prompting strategy was proposed by \citet {yang2023dual}, decomposing DST into slot and value generation tasks. Compared to the above research, our work complements the zero-shot prompting techniques to harness the capabilities of LLMs without collecting human-annotated data. 
% \shruti{specify how ours is similar and different from these}
\section{Methodology}
\label{sec:method}
% \shruti{go through} \\
% \andy{go through} \\
% \joel{DONE} \\

Figure~\ref{fig:pipeline} outlines the \method's data generation pipeline. It utilizes just the dialogue schema and a set of handcrafted templates to generate fluent dialogues with dialog state annotations. Specifically, each dialog generated using \method\ comprises a quartet of dialogue history, system turn, user turn, and the current turn's dialog states. \method\ employs a three-step approach for generating dialogue data. We explain each step below.
% Note that \method\ is designed to be extendable to new domains without much modifications. 
% Recent DST approaches necessitate the use of conversation history alongside the recent most system-user turns to predict the corresponding DST values. Traditionally, conversation history has been represented by concatenating all prior system-user utterances \cite{lee-etal-2021-dialogue , lin-etal-2021-leveraging, peng-etal-2021-soloist}. However, conversation histories can often become lengthy. This is especially problematic given the limitations on the prompt length of LLMs, posing challenges for incorporating multiple examples for in-context learning. Furthermore, much of the history may contain redundant or irrelevant information with respect to the current DST prediction. Thus, inspired by prior work \cite{hu-etal-2022-context, li-etal-2022-controllable}, we represent conversation history as the cumulative dialogues tate up to the current point.  On this theme, \method\ generates data samples as a quartet of \textit{history DST, current system turn, current user turn,} and \textit{current DST}.

\subsection{Dialogue Structure Synthesis}

\paragraph{Abstract Dialogue Model.}
The effectiveness of \method\ in generating meaningful dialogues relies on its strategic selection of system and user dialogue acts. A dialogue act, represented by intent and its associated slot-value pairs, indicates the specific communicative action of a user and the system \cite{core1997coding, Traum1999, budzianowski-etal-2018-multiwoz}. Selecting valid dialogue acts for each system and user turn is non-trivial, as random pairing may yield incoherent and illogical dialogues. To address this issue, we adopt an approach similar to that of \citet{campagna-etal-2020-zero} by creating an abstract dialogue model. We define it as a set of system-user dialogue acts along with their valid transitions. Table \ref{tab:abstract_model} depicts the system-user intents and their valid transitions used in the abstract dialogue model. Our meticulously curated list of system-user intent transitions is independent of any dialogue domains and datasets, following a natural dialogue progression. Hence, it proffers greater generalizability and scalability. 
% As an example, Figure \ref{fig:pipeline} illustrates the slots for the domain \textit{hotel}, such as \textit{area}, which can take values \{\textit{east, west, north, south, centre}. For instance, the dialogue act \textit{inform(domain=train, destination=Cambridge)} signifies the user's intent to provide information regarding the `train destination', specifying it as `Cambridge'. 

% We first randomly select a domain from the set of domains ((\textit{attraction, hotel, restaurant, taxi, train}) in MultiWOZ).

\paragraph{Synthesizing Dialogue Structure from Abstract Dialogue Model.} For synthesizing a sample,  we begin by selecting a system-user intent pair from Table \ref{tab:abstract_model}. Following previous works \cite{campagna-etal-2020-zero, hu-etal-2022-context}, we represented dialogue history as the accumulated dialogue state. The dialogue history is constructed by randomly selecting slot-value pairs from the dialogue schema\footnote{Dialogue schema presents a structured representation of the valid slots and values across different domains} following the chosen system intent. The system and user dialogue acts are sampled based on the dialogue history and the selected system-user intent. Lastly, we sample the current belief state values, considering the dialogue history and the user dialogue act. This approach guarantees the generation of coherent and contextually appropriate dialogue structure. 
% \joel{I didn't quite get what the vairable $n$ was meant to indicate here, and I'm not sure if it's used anywhere else?} \shruti{removed n}

\subsection{Template Response Generation}

\begin{table}[t!]
% \centering
\resizebox{\columnwidth}{!}{%
\begin{tabular}{l|l}
\hline
\textbf{System Intent} & \textbf{User Intent} \\ \hline
start & inform \\ \hline
inform & inform, update, reqmore, confirm, book \\ \hline
nooffer & update, recheck, end \\ \hline
select & pick, update, reqmore \\ \hline
recommend & select, update, reqmore \\ \hline
request & inform \\ \hline
booking-request & inform \\ \hline
booking-inform & book, nobook, update, reqmore, inform \\ \hline
offerbooked & new\_domain, confirm, end \\ \hline
booking-book & new\_domain, confirm, end \\ \hline
booking-nobook & new\_domain, recheck, end \\ \hline
\end{tabular}%
}
\caption{Coherent system-user intents. For each system intent, we define the list of user intents that can indicate a natural next turn flow. }
\label{tab:abstract_model}
\end{table}

Prompting LLMs to generate free-flowing dialogues from the raw conversation structure offers limited control over its characteristics. As highlighted by \citet{li-etal-2022-controllable} and \citet{chen-etal-2023-places}, unconstrained generation based solely on the dialogue history or topic can produce erroneous dialogues, often necessitating human review and correction. 
% Moreover, in the case of DST, it led to inconsistent data. At times the generated dialogues  are not grounded in the given belief states.  We present such erroneous dialogues amples in the Appendix \ref{sec:appendix}. While creating synthetic data, this posits a substantial bottleneck.
On the other hand, prompting an LLM to modify a skeletal dialogue offers better control. Thus, drawing inspiration from \citet{Rastogi_Zang_Sunkara_Gupta_Khaitan_2020} and \citet{kale-rastogi-2020-template}, we adopt the template-guided approach that enables fine-grained control over dialogue content. While \citet{kale-rastogi-2020-template} primarily offer templates for system response generation, their method entails crafting separate templates for each domain, dialogue act, and slot triplet. 
% Specifically, they map each $(d,a, s)$ to a template description $t_{i}$, resulting in nearly 233 templates, demanding extensive human effort.
This results in more than $200$ templates, demanding extensive human effort. Furthermore, these templates are not domain- and slot-agnostic, demanding effort with each new domain and schema modification. Building upon their methodology, we introduce domain-agnostic templates for both system and user responses.

\begin{table}[t!]
\centering
\resizebox{\columnwidth}{!}{%
\begin{tabular}{l|l}
\hline
\textbf{dialogue Act}   & \textbf{Template} \\ \hline
recommend (d, s, v)   & I would suggest the <d> with <s> <v> \\ \hline
offerbooked (d, s, v) & Booked <d> for <v> <s> \\ \hline
request (d, s, v)     & What is your preferred <d> <v> ? \\ \hline
\end{tabular}%
}
\caption{Selected system template responses}
\label{tab:templates_system}
\end{table}

\begin{table}[t!]
\centering
\resizebox{\columnwidth}{!}{%
\begin{tabular}{l|l}
\hline
\textbf{dialogue Act}   & \textbf{Template} \\ \hline
inform (d, s, v)   & The <d> <s> should be <v> \\ \hline
nobook (d, s, v) & No, don't book the <d> for <v> <s> \\ \hline
reqmore (d, s, v)     & What is the <d>'s <s> ? \\ \hline
\end{tabular}%
}
\caption{Selected user template responses}
\label{tab:templates_user}
\end{table}
Given a quartet of domain, dialogue act, slot, and value, respectively, we map it to a template that depends only on the dialogue act. Each template contains designated placeholder tokens for domain, slots, and values, which are substituted during template generation. This guarantees that the generated dialogues are grounded in the provided belief states. We utilize templates for just 22 dialogue acts (11 each for system and user), thus considerably reducing human efforts. We generate between 2 and 4 templates per dialogue act to encourage diversity. Also, our templates are domain-agnostic and can be scaled to newer domains without additional effort. Tables \ref{tab:templates_system} and \ref{tab:templates_user} illustrate some of our system and user templates, respectively. 

\subsection{LLM-based Template Modification} \label{sec:dialog_modification}

% \shruti{go through} \\
% \andy{go through} \\
% \joel{DONE} \\

While the templates offer natural language descriptions for both system and user responses, they lack linguistic and conversational variations. Additionally, as these templates are designed to be domain and slot-value agnostic, they may contain certain grammatical and fluency errors. As a result, transforming these template-based responses into more naturalistic and free-flowing language can lead to contextually appropriate dialogues. 
Following previous research efforts in synthetic data generation \cite{mehri-etal-2022-lad, xiang-etal-2022-asdot, li-etal-2022-controllable, chen-etal-2023-places}, we employ GPT-3.5 \cite{NEURIPS2020_1457c0d6} for converting the template responses to free-flowing dialogues. For this, we explore three distinct prompting strategies, detailed as follows.
% Previous research has demonstrated that LLMs perform well in data-to-text generation when provided with suitable prompts \cite{mehri-etal-2022-lad, xiang-etal-2022-asdot}. Therefore, we use the capabilities of \textit{gpt-3.5-turbo} to enhance the coherence of template-based utterances while maintaining their original meaning and alignment with DST values. Concretely, we experiment with three different approaches, outlined below.

We initially experimented with `\textit{dialogue-level prompting}', instructing the LLM to modify the entire two-turn dialogue. This approach led to a hallucination of slot-value pairs and the generation of disfluent dialogues as the LLM often merged or interchanged information between user and system utterances. We also encountered instances where one of the system-user responses was skipped, generating a single utterance. 
We then explore a `\textit{multi-step prompting}' approach, which employs a sequential prompting process. First, we prompt the LLM to refine the system template and then modify the user template independently by providing the modified system response. While this addresses the issue of skipped utterances, it still suffers from information blending between system and user responses, resulting in incorrect slot-value annotations and dialogues. 

To overcome these drawbacks, we opt for `\textit{utterance-level prompting}'. In this method, we refine the system and user template independently. This approach results in succinct responses strongly anchored in the template structure and consistent with the slot values. Importantly, it avoids the issue of information merging between system and user turns. We use this as our final prompting strategy. The prompt used is as follows:

\begin{figure}[h!]
 % \fboxsep 5pt % Adjust the padding size (e.g., 10pt)
  % \fboxrule 0.5pt % Adjust the border thickness (e.g., 1pt)
  \fbox{
    \parbox{0.95\columnwidth}{
      \textit{Following is a template <user/system> response for a conversation between a <domain> chatbot and a user. Paraphrase the template by making it more fluent, engaging, polite, and coherent. Also, correct grammatical mistakes. Reorder the sentences if necessary. \\ Strictly generate the response in the form of a JSON object \{`<user/system>\_paraphrased': ''\} with correct formatting (including curly brackets). Do not return anything else apart from the JSON object. \\ `<user/system>\_template': `<template>'}
    }
  }
  % \caption{Template refinement prompt.}
  \label{fig:refine_prompt}
\end{figure}

While the utterance modification using the above prompting scheme results in naturalistic conversations, we find that their stylistic diversity remains limited. Thus, to make the dataset more diverse, we use `\textit{paraphrase prompting}', to generate different paraphrased dialogue variants. Similar to \textit{utterance-level prompting}, we independently paraphrase both system and user responses to create the final dialogue. Our selection of prompts for paraphrasing is randomized from the following set:
% \joel{TODO: we seem to have used "dialog" and "dialogue " both. I think we ought to be consistent (probably doesn't matter which one though)}. 
% \joel{this needs to be formatted better. Andy might have some ideas here.}:

% \begin{itemize}
%     \item \textit{Rephrase the sentences while retaining the original meaning.}
%     \item \textit{Use synonyms or related words to express the sentences with the same meaning.}
%     \item \textit{Alter the style of the sentences.}
%     \item \textit{Change the structure of the sentences.}
%     \item \textit{Use conversational language and paraphrase the following sentences.}
%     \item \textit{Generate a crisp and to the point single sentence from the given sentences using conversational language.}
% \end{itemize}

\begin{figure}[h!]
 % \fboxsep 5pt % Adjust the padding size (e.g., 10pt)
  % \fboxrule 0.5pt % Adjust the border thickness (e.g., 1pt)
  \fbox{
    \parbox{0.95\columnwidth}{
     \textit{Rephrase the sentences while retaining the original meaning.} \\
    \textit{Use synonyms or related words to express the sentences with the same meaning.} \\
    % \item \textit{Alter the style of the sentences.}
    % \item \textit{Change the structure of the sentences.}
    \textit{Use conversational language and paraphrase the following sentences.} \\
    \textit{Generate a crisp and to the point single sentence from the given sentences using conversational language.}
    }
  }
  % \caption{Utterance paraphrasing prompt.}
  \label{fig:paraphrase_prompt}
\end{figure}

\begin{table*}[!ht]
\begin{center}
\resizebox{0.7\textwidth}{!}{%
\begin{tabular}{clccccccc}
\toprule
\multicolumn{1}{c}{\multirow{2}{*}{\textbf{Percentage}}} & \multicolumn{1}{c}{\multirow{2}{*}{\textbf{Method}}} & \multicolumn{7}{c}{\textbf{MultiWoZ 2.1}} \\ 
\cmidrule(lr){3-9}
& & \textbf{Attraction} & \textbf{Hotel} & \textbf{Restaurant} & \textbf{Taxi} & \textbf{Train} & \textbf{JGA$_{\mathrm{D}}$}  & \textbf{JGA$_{\mathrm{A}}$}\\
\midrule
\multicolumn{1}{c}{\multirow{2}{*}{$-$}} & Zero Shot & $71.8_{0.0}$ & $45.3_{0.2}$ & $63.1_{0.8}$ & $72.7_{0.6}$ & $61.5_{0.8}$ & $62.9_{0.3}$ & $39.9_{0.3}$\\
& Few Shot$_{\mathrm{random}}$ & $74.4_{0.2}$ & $48.8_{2.9}$ & $60.9_{5.3}$ & $74.0_{0.7}$ & $60.3_{2.1}$ & $63.7_{1.9}$ & $40.3_{2.4}$ \\
\cmidrule(lr){1-9}
\multicolumn{1}{c}{\multirow{2}{*}{$-$}} & Few Shot$_{\mathrm{unique_{all}}}$ & $72.0_{0.4}$ & $51.2_{1.8}$ & $65.3_{0.7}$ & $75.5_{0.7}$ & $69.0_{1.0}$ & $66.6_{0.3}$ & $45.3_{0.5}$ \\
& Few Shot$_{\mathrm{unique_{all}}_{5\mathrm{x}}}$  & $72.3_{0.6}$ & $51.6_{0.7}$ & $65.9_{1.3}$ & $74.6_{1.0}$ & $69.0_{0.6}$ & $66.7_{0.1}$ & $45.0_{0.1}$ \\
\cmidrule(lr){1-9}
\multicolumn{1}{c}{\multirow{2}{*}{$1\%$}} & Few Shot$_{\mathrm{\method}}$ & $72.6_{0.4}$ & $51.9_{0.4}$ & $66.9_{0.6}$ & $75.1_{0.1}$ & $\boldsymbol{68.7_{1.6}}$ & $67.1_{0.3}$ & $\boldsymbol{45.8_{0.3}}$ \\
& Few Shot$_{\mathrm{train}}$ & $73.9_{0.4}$ & $52.4_{0.6}$ & $67.3_{1.0}$ & $76.6_{0.5}$ & $66.0_{0.9}$ & $67.2_{0.3}$ & $45.0_{0.4}$ \\
\cmidrule(lr){1-9}
\multicolumn{1}{c}{\multirow{2}{*}{$5\%$}} & Few Shot$_{\mathrm{\method}}$ & $71.0_{0.9}$ & $52.1_{1.3}$ & $65.9_{1.5}$ & $76.3_{0.5}$ & $\boldsymbol{68.4_{0.3}}$ & $66.7_{0.6}$ & $44.9_{0.8}$\\
& Few Shot$_{\mathrm{train}}$  &$74.3_{1.0}$ & $54.2_{0.7}$ & $69.0_{1.6}$ & $78.6_{1.1}$ & $66.7_{0.9}$ & $68.6_{0.8}$ & $46.2_{1.1}$ \\
\cmidrule(lr){1-9}
\multicolumn{1}{c}{\multirow{2}{*}{$10\%$}} & Few Shot$^{\dagger}$$_{\mathrm{\method}}$  & $71.2_{0.9}$ & $51.5_{0.6}$ & $67.2_{1.5}$ & $76.3_{0.4}$ & $\boldsymbol{69.0_{0.3}}$ & $67.1_{0.2}$ & $45.4_{0.6}$\\
& Few Shot$_{\mathrm{train}}$  & $74.2_{0.2}$ & $53.8_{0.4}$ & $69.1_{1.3}$ & $78.3_{1.5}$ & $66.4_{0.9}$ & $68.3_{0.5}$ & $46.1_{0.8}$\\
\cmidrule(lr){1-9}
$100\%$ & Few Shot$^\dagger$$_{\mathrm{train}}$ & $74.0_{0.1}$ & $51.9_{0.3}$ & $69.0_{0.4}$ & $79.6_{0.4}$ & $70.4_{0.8}$ & $69.0_{0.0}$ & $46.0_{0.1}$\\
\midrule
\multicolumn{2}{c}{$\Delta_{\mathrm{\method}^\dagger - \mathrm{zero shot}}$} & $\downarrow0.6$ & $\uparrow6.2$ & $\uparrow4.1$ & $\uparrow3.5$ & $\uparrow7.5$ & $\uparrow4.2$  & $\uparrow5.5$\\
\multicolumn{2}{c}{$\Delta_{\mathrm{\method}^\dagger - \mathrm{random}}$}  & $\downarrow3.2$ & $\uparrow2.7$ & $\uparrow6.3$ & $\uparrow2.3$ & $\uparrow8.7$ & $\uparrow3.4$  & $\uparrow5.1$\\
\multicolumn{2}{c}{$\Delta_{\mathrm{\method}^\dagger / \mathrm{train}^\dagger}$} & $96.2$ & $99.2$ & $97.4$ & $95.8$ & $98.0$ & $97.4$ & $98.7$\\
\midrule
\multicolumn{1}{c}{\multirow{2}{*}{\textbf{Percentage}}} & \multicolumn{1}{c}{\multirow{2}{*}{\textbf{Method}}} & \multicolumn{7}{c}{\textbf{MultiWoZ 2.4}} \\ 
\cmidrule(lr){3-9}
& & \textbf{Attraction} & \textbf{Hotel} & \textbf{Restaurant} & \textbf{Taxi} & \textbf{Train} & \textbf{JGA$_{\mathrm{D}}$}  & \textbf{JGA$_{\mathrm{A}}$}\\
\midrule
\multicolumn{1}{c}{\multirow{2}{*}{$-$}} & Zero Shot & $78.2_{0.2}$ & $52.1_{0.1}$ & $67.2_{0.7}$ & $72.6_{0.5}$ & $66.1_{0.1}$ & $67.2_{0.2}$ & $45.6_{0.3}$\\
& Few Shot$_{\mathrm{random}}$ & $81.3_{0.6}$ & $51.6_{4.3}$ & $63.2_{6.1}$ & $73.6_{0.4}$ & $63.2_{2.5}$ & $66.6_{1.9}$ & $44.2_{2.6}$ \\
\cmidrule(lr){1-9}
\multicolumn{1}{c}{\multirow{2}{*}{$-$}} & Few Shot$_{\mathrm{unique_{all}}}$ & $79.1_{0.8}$ & $56.8_{1.5}$ & $66.9_{1.4}$ & $76.4_{0.4}$ & $73.2_{0.4}$ & $70.5_{0.6}$ & $50.4_{1.0}$ \\
& Few Shot$_{\mathrm{unique_{all}}_{5\mathrm{x}}}$ & $78.7_{0.2}$ & $57.4_{0.9}$ & $67.6_{0.5}$ & $74.8_{0.1}$ & $73.8_{0.8}$ & $70.4_{0.2}$ & $50.4_{0.4}$ \\
\cmidrule(lr){1-9}
\multicolumn{1}{c}{\multirow{2}{*}{$1\%$}} & Few Shot$_{\mathrm{\method}}$ & $79.4_{0.5}$ & $57.2_{0.8}$ & $69.2_{0.3}$ & $76.2_{0.4}$ & $\boldsymbol{72.5_{1.6}}$ & $70.9_{0.5}$ & $51.0_{1.1}$ \\
& Few Shot$_{\mathrm{train}}$ & $81.4_{0.3}$ & $58.8_{2.4}$ & $72.1_{2.3}$ & $77.1_{0.2}$ & $70.2_{2.2}$ & $71.9_{0.5}$ & $52.1_{1.0}$\\
\cmidrule(lr){1-9}
\multicolumn{1}{c}{\multirow{2}{*}{$5\%$}} & Few Shot$_{\mathrm{\method}}$ & $79.1_{1.4}$ & $56.8_{1.1}$ & $69.8_{1.6}$ & $77.1_{0.9}$ & $\boldsymbol{72.4_{2.0}}$ & $71.1_{1.2}$ & $50.4_{1.8}$\\
& Few Shot$_{\mathrm{train}}$ & $81.3_{0.6}$ & $60.0_{0.4}$ & $74.5_{0.9}$ & $78.5_{0.6}$ & $72.4_{1.9}$ & $73.4_{0.5}$ & $54.2_{1.0}$\\
\cmidrule(lr){1-9}
\multicolumn{1}{c}{\multirow{2}{*}{$10\%$}} & Few Shot$^{\dagger}$$_{\mathrm{\method}}$ & $77.9_{0.5}$ & $57.6_{0.3}$ & $69.9_{0.5}$ & $77.1_{0.6}$ & $\boldsymbol{73.2_{0.8}}$ & $71.1_{0.2}$ & $50.9_{0.3}$\\
& Few Shot$_{\mathrm{train}}$ & $82.1_{0.9}$ & $60.6_{0.8}$ & $75.0_{0.8}$ & $79.1_{0.9}$ & $71.3_{1.2}$ & $73.6_{0.3}$ & $53.8_{0.9}$\\
\cmidrule(lr){1-9}
$100\%$ & Few Shot$^\dagger$$_{\mathrm{train}}$ & $84.0_{0.4}$ & $60.0_{0.4}$ & $75.9_{0.4}$ & $81.3_{0.2}$ & $74.7_{0.4}$ & $75.2_{0.1}$ & $55.2_{0.2}$\\
\midrule
\multicolumn{2}{c}{$\Delta_{\mathrm{\method}^\dagger - \mathrm{zero shot}}$} & $\downarrow0.3$ & $\uparrow5.5$  & $\uparrow2.7$ & $\uparrow4.5$ & $\uparrow7.1$ & $\uparrow3.9$ & $\uparrow5.3$\\
\multicolumn{2}{c}{$\Delta_{\mathrm{\method}^\dagger - \mathrm{random}}$} & $\downarrow3.4$  & $\uparrow6.0$ & $\uparrow6.7$ & $\uparrow3.5$ & $\uparrow10.1$ & $\uparrow4.5$ & $\uparrow6.7$\\
\multicolumn{2}{c}{$\Delta_{\mathrm{{\method}}^\dagger / \mathrm{train}^\dagger}$}  & $92.7$ & $96.0$ & $92.1$ & $94.8$ & $98.0$ & $94.5$ & $92.2$\\
\bottomrule
\end{tabular}%
}
\end{center}
\caption{Comparison of per-domain Joint Goal Accuracy (JGA$_{\mathrm{D}}$) and all-domain Joint Goal Accuracy (JGA$_{\mathrm{A}}$) on MultiWoZ 2.1 and 2.4 using zero-shot, random few-shot, and retrieval-based few shot prompting with different percentages of synthetic and training data.}
\label{tab:results_main}
\end{table*}
\section{Experimental Setup}

% \shruti{modify \ go through} \\
% \andy{modify \ go through} \\
% \joel{DONE} \\

\subsection{Synthetic Data Generation}
\label{sec:data_gen}
Using \method, we create two types of synthetic corpora. In line with prior DST works \cite{wu-etal-2019-transferable, hu-etal-2022-context}, we generate data equivalent to $1\%$, $5\%$, and $10\%$ of the training data size, ensuring a fair comparison with regard to number of samples in the retrieval bank. Each set contains $1)$ $50\%$ of conversations featuring new slot-value pairs, $2)$ $15\%$ of conversations with no new belief states introduced $3)$ $10\%$ each of conversation starters and terminators, $4)$ $10\%$ of conversations updating existing slots with new values, and  $5)$ $5\%$ involving the repetition or deletion of prior slot-value information.
% These datasets are crafted to emulate a realistic distribution of conversations while encompassing diverse dialogue flows.
% We find that the MultiWOZ training corpus follows a similar trend. 
This ensures that the data bank follows a realistic distribution of conversations while encompassing diverse dialogue flows. Moreover, such careful data curation is known to stabilize ICL performance \cite{chang-jia-2023-data}. Additionally, we generate sampling invariant versions of synthetic datasets, denoted as $\mathrm{unique_{all}}$ and $\mathrm{unique_{all}}_{5\mathrm{x}}$. The $\mathrm{unique_{all}}$ dataset includes all valid unique dialogue flows, while$\mathrm{unique_{all}}_{5\mathrm{x}}$ includes five instances of each unique dialogue flow. This results in a dataset of about $7$k and $25$k dialogues, respectively. Detailed information regarding these datasets can be found in Appendix \ref{appendix:data_gen}.

\subsection{In-Context Learning Model}
Our experiments are based on the IC-DST framework introduced by \citet{hu-etal-2022-context}. It reformulates DST as a text-to-SQL problem, using a tabular description of the ontology followed by relevant in-context examples in the LLM prompt. The IC-DST framework leverages the \textit{text-davinci-codex} version \cite{chen2021evaluating} of OpenAI's GPT-3 model \cite{NEURIPS2020_1457c0d6}. It uses the cumulative dialogue state to represent conversation history. This design choice enhances efficiency, incorporates more in-context examples, and performs effectively in the presence of domain shifts. Additionally, IC-DST introduces a novel similarity score to retrieve better in-context examples. We encourage readers to refer to \citet{hu-etal-2022-context} for a comprehensive understanding of the IC-DST framework.

We introduce specific modifications to the IC-DST framework, reducing its complexities and making it suitable for current versions of GPT models. Firstly, due to the deprecation of the \textit{text-davinci-codex}, we experiment with \textit{gpt-3.5-turbo}, a newer chat model that exhibits similar coding capabilities. Secondly, the IC-DST framework uses explicit fine-tuning of the retriever on the training data. This process needs compute resources and time and presupposes access to training data. Consequently, we have adopted an off-the-shelf solution in the form of Sentence-BERT \cite{reimers-gurevych-2019-sentence}, specifically the \textit{all-mpnet-base-v2} model \cite{NEURIPS2020_c3a690be}. We keep the rest of the formulations unchanged. 
% This choice removed any need for fine tuning from the entire system, where the retriever operates without any prior knowledge of the DST examples, enabling on-the-fly usage. 
% \joel{nice}

\subsection{Dataset}

\paragraph{MultiWOZ 2.1 \cite{eric-etal-2020-multiwoz}}is a multi-domain human-to-human dialogue dataset that contains over $10$K dialogues across $8$ domains. This is the updated version of the original MultiWOZ 2.0 dataset \cite{budzianowski-etal-2018-multiwoz}. MultiWOZ 2.1 is a widely used benchmark for DST and in dialogue systems research.

\paragraph{MultiWOZ 2.4 \cite{ye-etal-2022-multiwoz}}builds on top of the 2.1 version and makes substantial changes to the validation and test sets. MultiWOZ 2.4 can be viewed as a cleaner version of MultiWOZ 2.1 that better reflects model performance.

% \joel{TODO: missing cites for both of these}

% \subsection{Baselines}

\subsection{Evaluation Metrics}

% \shruti{modify \ go through} \\
% \andy{modify \ go through} \\
% \joel{DONE} \\

We employ the conventional Joint Goal Accuracy (JGA) as our evaluation metric. This metric considers a prediction correct when all slots-values match the ground truth. We report the \textit{All-Domain Joint Goal Accuracy (JGA$_{\mathrm{A}}$)} for the overall performance and the \textit{Per-domain Joint Goal Accuracy (JGA$_{\mathrm{D}}$)} for domain-level performance \cite{wu-etal-2019-transferable, hu-etal-2022-context}. 
% Similar to \citet{hu-etal-2022-context}, we use state changes as a label for prediction. 
% We report the Average Goal Accuracy (AGA) \cite{Rastogi_Zang_Sunkara_Gupta_Khaitan_2020} and Flexible Goal Accuracy (FGA) \cite{dey-etal-2022-towards} in Appendix \ref{appendix:extra_results}.

\section{Results and Discussion}

Table \ref{tab:results_main} presents our results for MultiWOZ 2.1 and 2.4. The zero-shot setting is the only baseline that does not rely on any human-annotated data, similar to our approach. We also report on a $\mathrm{random}$ setting, where we randomly add $2$ examples per domain (resulting in $10$ examples) from the training data to form a static set of in-context examples. Additionally, we assess the performance of synthetic and training data at different percentages as explained in section \ref{sec:data_gen}. For all setups, the average performance over $3$ runs is reported.

% \begin{table}[t!]
% \begin{center}
% \resizebox{\columnwidth}{!}{%
% \begin{tabular}{lcc}
% \toprule
% \textbf{Method} & \textbf{MultiWoZ 2.1} & \textbf{MultiWoZ 2.4} \\ \midrule
% Zero Shot & $39.9_{0.3}$ & $45.6_{0.3}$ \\
% Few Shot$_{\mathrm{random}}$ & $40.3_{2.4}$ & $44.2_{2.6}$ \\
% Few Shot$_{\mathrm{unique_1}}$ & $45.3_{0.5}$ & $50.4_{1.0}$ \\
% Few Shot$_{\mathrm{unique_5}}$ & $45.0_{0.1}$ & $50.4_{0.4}$ \\
% Few Shot$_{\mathrm{train}}$ & $46.0_{0.1}$ & $55.2_{0.2}$ \\
% \bottomrule
% \end{tabular}%
% }
% \end{center}
% \caption{Comparison of combined Joint Goal Accuracy (JGA) on MultiWoZ 2.1 and 2.4 using zero, random, synthetic, and training data.}
% \label{tab:result_unique}
% \end{table}

\begin{table}[t!]
\begin{center}
\resizebox{\columnwidth}{!}{%
\begin{tabular}{lccccccc}
\toprule
\textbf{Method} & \textbf{Attraction} & \textbf{Hotel} & \textbf{Restaurant} & \textbf{Taxi} & \textbf{Train} & \textbf{JGA$_{\mathrm{D}}$}  & \textbf{JGA$_{\mathrm{A}}$}\\ 
\midrule
Few Shot$_{\mathrm{T}}$ & $73.3_{0.7}$ & $49.7_{0.2}$ & $63.6_{1.4}$ & $75.9_{0.5}$ & $66.7_{0.6}$ & $65.8_{0.3}$ & $43.8_{0.2}$ \\
% Few Shot$_{\mathrm{N}}$  & $72.7_{0.6}$ & $51.4_{0.9}$ & $64.8_{1.8}$ & $76.1_{1.1}$ & $69.7_{1.2}$ & $66.9_{0.7}$ & $45.5_{1.0}$ \\
Few Shot$_{\mathrm{LLM}}$  & $72.6_{0.4}$ & $51.9_{0.4}$ & $66.9_{0.6}$ & $75.1_{0.1}$ & $68.7_{1.6}$ & $67.1_{0.3}$ & $45.8_{0.3}$ \\
\bottomrule
\end{tabular}%
}
\end{center}
\caption{Ablation study of \method. Few Shot$_\mathrm{T}$ and Few Shot$_\mathrm{LLM}$ refer to template and LLM-modified data, respectively.}
\vspace{-3mm}
\label{tab:result_ablation}
\end{table}

\paragraph{Synthetic Data Consistently Beats Zero-Shot.} Few-shot prompting using data generated from {\method} and the zero-shot setups illustrate scenarios where no training data is used. This is particularly relevant to practical settings where obtaining human-labeled data can be prohibitively expensive in terms of cost and human effort. From table \ref{tab:results_main}, observe that few-shot using data from {\method} leads to substantial gains over zero-shot. Specifically, we observe about $4\%$ and $5\%$ improvements for JGA$_{\mathrm{D}}$ and JGA$_{\mathrm{A}}$, respectively, across both the datasets. Moreover, it gives notably high gains on the two worst performing domains, about $6\%$  on \textit{hotel} and $7\%$ on \textit{train}. In summary, synthetic data may provide a good solution when no training data is available.

\paragraph{Retrieval-Based ICL with Synthetic Data Outperforms ICL with Random Training Examples.} In some scenarios, ML practitioners may have access to a limited number of in-domain examples. Therefore, using a few static examples for few-shot learning is a relevant baseline. Table \ref{tab:results_main} reveals that utilizing randomly selected in-domain examples leads to similar or worse performance than the zero-shot setting. Notably, the performance drops significantly on \textit{restaurant} and \textit{train} domains. This observation aligns with previous findings \cite{liu-etal-2022-makes}, highlighting the high variance in results and emphasizing that random example selection is not an effective choice for ICL. {\method} offers improvements of approximately $5-6\%$ on the JGA$_{\mathrm{D}}$ and JGA$_{\mathrm{A}}$ for both MultiWOZ versions across most domains. Interestingly, we notice substantial gains in the \textit{attraction} domain. We conjecture that these gains can be attributed to the distribution in the test split. We discuss more on this in Appendix \ref{appendix:discussion}.

\paragraph{{\method} Competes Effectively with Training Data.} Table \ref{tab:results_main} reports the performance on different percentage splits of training data. 
% This presents an upper bound on the synthetic data performance. 
The results indicate that {\method} consistently recovers over $95\%$ and $92\%$ of the training data performance on MultiWOZ 2.1 and 2.4 across all domains. Surprisingly, it even outperforms the $1\%$ training data setup in MultiWOZ 2.1. Also, there are improvements of $1-3\%$ on the \textit{train} domain for both versions. Moreover, it significantly reduces the performance gap, particularly in the \textit{hotel} domain, which exhibited the poorest performance in the zero-shot setting. 
% In conclusion, we advocate that \method\ produces high-quality synthetic DST data, serving as a reliable surrogate for training data in the in-context learning paradigm.

% \paragraph{Quality Trumps Quantity in Synthetic Data.}  In Section \ref{sec:data_gen}, we emphasize the importance of meticulously curating the ICL data pool for improved few-shot learning. By comparing the outcomes in Tables \ref{tab:results_main} and 

\paragraph{Quality Trumps Quantity in Synthetic Data.}  In Section \ref{sec:data_gen}, we emphasize the importance of meticulously curating the ICL data pool for improved few-shot learning. From Table \ref{tab:results_main}
% and \ref{tab:result_unique}
it becomes evident that few-shot learning with $\mathrm{unique_{all}}$ and $\mathrm{unique_{all}}_{5\mathrm{x}}$ data almost never surpasses the performance of the carefully curated data. Despite $\mathrm{unique_{all}}$ and $\mathrm{unique_{all}}_{5\mathrm{x}}$ being approximately $14$x and $47$x times larger than the $1\%$ data subset, respectively, it is clear that having a substantial representation of relevant examples is superior to having an equal representation of all examples. Moreover, less relevant examples can introduce noise and adversely affect predictions if the proportions of labels appearing in context differ greatly from the test instance \cite{pmlr-v139-zhao21c}. Nevertheless, we still maintain a consistent improvement of over $5\%$ compared to the zero-shot and random settings, underscoring the effectiveness of our synthetic data.

\paragraph{Template Data or LLM  Modified Data?} Table \ref{tab:result_ablation} presents an ablation study conducted on the $1\%$ split of MultiWoZ 2.1. We observe that relying solely on template data yields improved performance in the \textit{attraction} domain but significantly lower results in the \textit{hotel}, \textit{restaurant}, and \textit{train} domains, resulting in an overall decrease in performance. Transitioning from templates to more naturalistic conversations leads to an approximate $2\%$ improvement on JGA$_{\mathrm{D}}$ and JGA$_{\mathrm{A}}$. There is also a noticeable improvement in the \textit{restaurant}, \textit{hotel}, and \textit{train} domain. Comparing these findings with Table \ref{tab:results_main}, we observe that relying solely on template data results in an improvement of nearly $4\%$ in JGA$_{\mathrm{A}}$. Therefore, even without LLMs, \method\ offers significant gains over the zero-shot setting.

\paragraph{Synthetic Data Helps Unveil Annotation Bias.}
% \atharva{TODO}
Inconsistent annotation has been a pervasive issue in DST datasets \cite{zang-etal-2020-multiwoz, han2021multiwoz, ye-etal-2022-multiwoz}. While MultiWoZ 2.4 presents a much cleaner version, our study uncovers a distinct concern unanswered previously: incongruities related to domain ontology. More precisely, our examination has revealed that the current annotations treat parking and internet slots labeled as `yes' as synonymous with `free.' However, these are two separate slot values in the schema and convey distinct meanings. To illustrate, when parking slots are marked as 'yes,' it generally indicates the availability of parking. Nevertheless, it does not necessarily imply that the parking is free; users might still be required to pay for parking despite the availability of slots.

\begin{table*}[!ht]
\begin{center}
\setlength{\tabcolsep}{-12pt} 
\resizebox{0.9\textwidth}{!}{%
\begin{tabular}{cc@{}cccccccc}
\toprule
\multicolumn{1}{c}{\multirow{3}{*}{\textbf{Percentage}}} & \multicolumn{4}{c}{\textbf{Utterance Modification}} & \multicolumn{4}{c}{\textbf{Utterance Paraphrasing}} &  \multicolumn{1}{p{2cm}}{\centering \textbf{Cost}} \\
\cmidrule(lr){2-5}
\cmidrule(lr){6-9}
\multicolumn{1}{p{4cm}} . & \multicolumn{1}{p{2.5cm}}{\centering \textbf{Avg sys.}} & \multicolumn{1}{p{2.5cm}}{\centering \textbf{Avg sys.}} & \multicolumn{1}{p{2.5cm}}{\centering \textbf{Avg user}} & \multicolumn{1}{p{2.5cm}}{\centering \textbf{Avg user}} & \multicolumn{1}{p{2.5cm}}{\centering \textbf{Avg sys.}} & \multicolumn{1}{p{2.5cm}}{\centering \textbf{Avg sys.}} & \multicolumn{1}{p{2.5cm}}{\centering \textbf{Avg user}} & \multicolumn{1}{p{2.5cm}}{\centering \textbf{Avg user}} & \multicolumn{1}{p{2cm}}{\centering \textbf{(USD)}}\\
& \multicolumn{1}{p{2.5cm}}{\centering \textbf{inp. tok.}} & \multicolumn{1}{p{2.5cm}}{\centering \textbf{ out. tok.}} & \multicolumn{1}{p{2.5cm}}{\centering \textbf{inp. tok.}} & \multicolumn{1}{p{2.5cm}}{\centering \textbf{out. tok.}} & \multicolumn{1}{p{2.5cm}}{\centering \textbf{inp. tok.}} & \multicolumn{1}{p{2.5cm}}{\centering \textbf{out. tok.}} & \multicolumn{1}{p{2.5cm}}{\centering \textbf{inp. tok.}} & \multicolumn{1}{p{2.5cm}}{\centering \textbf{out. tok.}} & \\

\midrule
$1\%$ ($\approx549$ data) & $120.46$ & $28.93$ & $114.02$ & $25.63$ & $41.09$ & $30.15$ & $37.98$ & $26.90$ & $\$0.38$ \\
$5\%$ ($\approx2748$ data) & $119.54$ & $27.95$ & $114.27$ & $25.78$ & $40.23$ & $29.52$ & $37.83$ & $26.46$ & $\$1.88$ \\
$10\%$ ($\approx5495$ data) & $119.95$ & $28.23$ & $114.14$ & $25.91$ & $40.37$ & $29.41$ & $38.06$ & $26.54$ & $\$3.78$ \\

\bottomrule
\end{tabular}%
}
\end{center}
\caption{\refine{Cost Analysis of \method\ in USD. Leveraging OpenAI's GPT-3.5-turbo, the expense is $\$0.0010$ per $1000$ input tokens and $\$0.0020$ per $1000$ output tokens. With these cost projections, generating a synthetic dataset equivalent in size to MultiWoZ ($\approx55$k examples) using \method\ will cost less than $\$40$!}}
\label{tab:cost_analysis}
\end{table*}

\begin{figure}[t!]
    \centering
    \includegraphics[width=0.8\columnwidth]{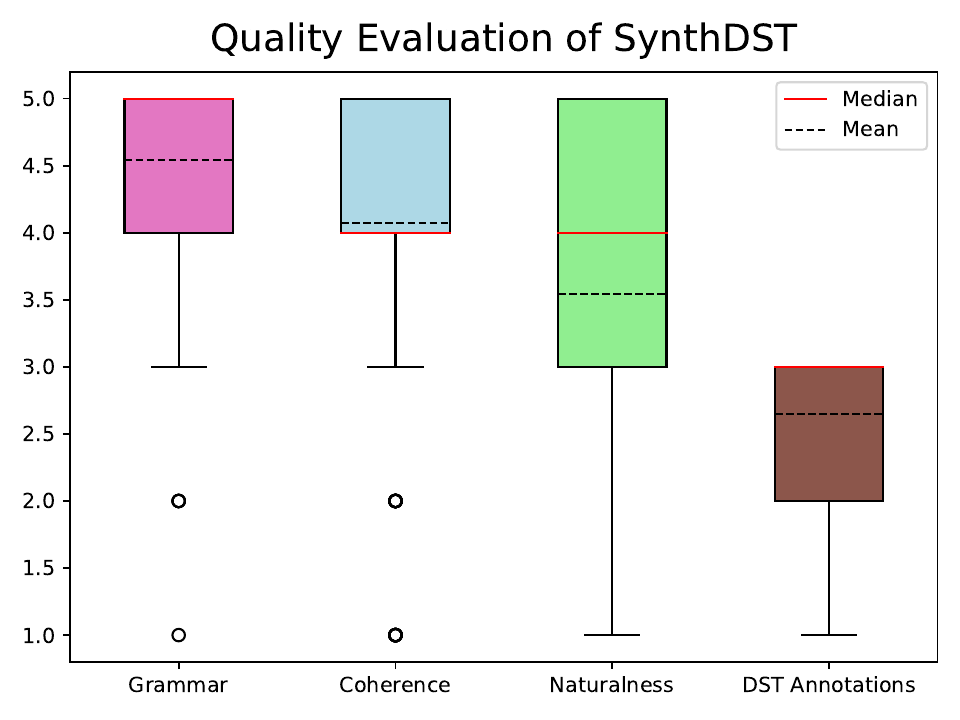}
    \caption{Box plot of Human evaluation scores.}
    \label{fig:human_eval}
\end{figure}

\section{Dataset Quality Analysis}

\paragraph{Is the data generated by \method\ of Good Quality?}
As \method\ is a human-involvement-free synthetic data generation approach (except for template definition), assessing its quality is crucial. Consequently, we conducted a human evaluation on 200 dialogues from our $1\%$ dataset split. Four evaluators, experienced in dialogue systems research, evaluate the data. Given the generated samples containing the dialogue history, average system utterance, average user utterance, and new dialogue state, the evaluators assessed the dialogues on four dimensions, namely, \textbf{\textit{Grammar}}, \textbf{\textit{Coherence}}, \textbf{\textit{Naturalness}}, and \textbf{\textit{Annotations}}. The annotations are rated from 1-3, whereas the others are graded on a 1-5 scale. The detailed scales are given in Appendix \ref{appendix:human_eval}. 

In Figure \ref{fig:human_eval}, we present the results of our human evaluation. The majority of the dataset demonstrates high scores for \textbf{\textit{Grammar}}, indicating grammatical correctness and minimal mistakes. For \textbf{\textit{Coherency}}, both the mean and median scores exceed $4$, signifying that the dialogues are mostly coherent and logically structured. While \textbf{\textit{Naturalness}} exhibits slightly more variability, the mean, and median still surpass 4, indicating that most dialogues maintain a natural conversational flow resembling real-world conversations. Lastly, the \textbf{\textit{Annotations}} scale attains a median of $3$ and a mean > $2.5$, suggesting that most of the annotations are correct.

\paragraph{Is \method\ More Cost-Effective than Human Annotation?}
% \shruti{TODO}
Creating the MultiWOZ dataset involved 1,249 workers and incurred a cost of approximately \$$30,000$, excluding post-processing expenses \cite{budzianowski-etal-2018-multiwoz, li-etal-2022-controllable}. In contrast, \method\ significantly reduces both cost and time requirements. \refine{Specifically, \method\ utilizes a total of $4$ OpenAI API calls for each sample, $1$ for modifying the system template into an utterance, 1 for modifying the user template into an utterance, then 1 for further paraphrasing the system utterance, and lastly for paraphrasing the user utterance. Table \ref{tab:cost_analysis} presents the details of input-output tokens utilization and the total cost for each prompting step across different data splits. We see that \method\ can generate an entire MultiWOZ-sized dataset ($\approx 55$k dialogues) in just about $\$40$. Moreover, generating $1\%$ equivalent data requires less than $\$1$ while maintaining the DST performance. Thus, \method\ presents a cost-effective method to collect DST data. }

% Assuming a minimum hourly wage of $8$\$, the whole process would take up to 3,750 work hours \cite{li-etal-2022-controllable}.

% Synthetic data is definitely at a gain compared to human annotations for time efficiency. It is only limited by the per token latency of LLMs, which is much lower compared to getting the data annotated manually. 

\section{Conclusion and Future Work}
% \shruti{TODO} \\
% \atharva{TODO} \\
% We present a generalised synthetic dialogue generation framework {\method} that only needs dialogue schema. It uses template-guided LLM-based approach to generate dialogues with diverse dialogue state changes. This framework allows the use of in-context learning for dialogue state tracking in the absence of training data. Performance with the {\method} reaches close to the performance with training data on this task. This opens the possibility to support new domains without the need for cumbersome and expensive training data collection, and also reduces some human bias from these datasets. Note that our approach experiments only with MultiWOZ domains and expeimenting with more diverse domains is a good direction for future resaearch. Also, we do not impose guardrails to check hallucinations in the LLM modified data.

In this work, we present {\method}, a synthetic data generation framework that leverages the dialogue schema to create coherent dialogues with DST annotations using a template-guided LLM-based approach. This framework enables the use of in-context learning for DST without human-annotated training data. Performance with {\method} reaches close to the performance with training data on dialogue state tracking. This opens the possibility of supporting new domains without needing cumbersome and expensive training data collection. Moreover, it also reduces some annotation bias from these datasets. 

\refine{Numerous potential avenues for future research emerge from our current work. While we experiment only with the MultiWOZ datasets, \method\ can readily be extended to other corpora. While \method\ predominantly relies on the close-sourced OpenAI GPT-3 model, it would be interesting to see how it performs with open-sourced LLMs. We encourage further research that validates its performance across diverse domains and models.} Moreover, our approach does not incorporate safeguards to detect hallucinations in LLM-generated data, which is a direction for future investigations.
\section{Limitations}
\refine{We designed {\method} as a domain-agnostic framework to enable scalability across different domains. However, this domain-agnostic approach comes with a trade-off -- it struggles to capture inter-slot dependencies. For instance, when the slot \textit{"attraction-type"} contains \textit{"sports,"} it should ideally retrieve sports-related attractions for the \textit{"attraction-name"} slot. Unfortunately, the current framework cannot accomplish this without compromising its domain-agnostic nature. Furthermore, {\method} lacks a post-hoc human correction module, resulting in the retention of such potentially erroneous examples in the dataset. Nevertheless, such examples are few and far between, as indicated by the high human evaluation scores. Thus, it's important to emphasize that despite these challenges, {\method} continues to deliver commendable performance.}
\section{Ethical Consideration}
This work uses LLMs for synthetic data generation. It makes an effort to ensure grounded and consistent data is generated by the LLM, however there can still be hallucinations and/or inconsistencies in the predictions. It is highly recommended to implement further guardrails to use such data synthesis approaches in real world scenarios.
% \andy{TODO} \\
% \joel{TODO} \\
\section{Acknowledgements}

The authors would like to thank Jeffrey Nichols, Stephen Pulman, Barry Theobald, Nidhi Rajshree, and the anonymous reviewers for their help and feedback.
Coo

% add acknowledgemens for final paper
% \input{sections/acknowledgements}

% \newpage
\bibliography{anthology,custom}

\newpage
\appendix

\section{Appendix}
\label{sec:appendix}
% \subsection{Performance wrt Other Evaluation Metrics}
% \label{appendix:extra_results}

\subsection{Synthetic Data Generation}
\label{appendix:data_gen}

Table \ref{tab:synthDST_dist} contains the domain distribution of the different splits of \method. For the $1\%$, $5\%$, and $10\%$ percentage data, we uniformly sample each domain data according to the sampling scheme explained in Section \ref{sec:data_gen}. For the $\mathrm{synthetic_1}$ and $\mathrm{synthetic_5}$ datasets, we observe an uneven distribution of domains. This disparity arises due to our emphasis on acquiring unique system-user dialogue act pairs. Since each domain has a distinct number of dialogue acts, the distribution becomes skewed.

\begin{figure*}[t!]
    \centering
    \includegraphics[width=\textwidth]{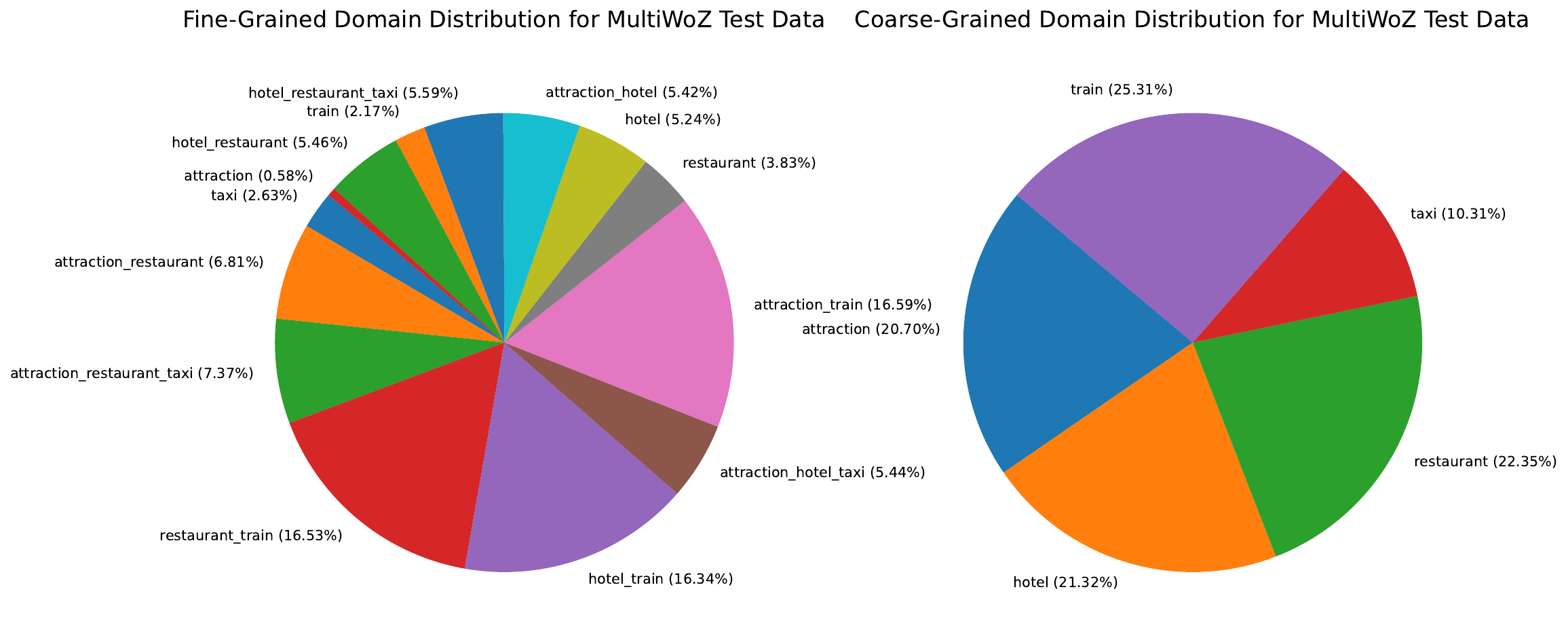}
    \caption{Domain distribution for MultiWoZ test data.}
    \label{fig:domain_dist}
\end{figure*}

\begin{table}[h!]
\begin{center}
\resizebox{\columnwidth}{!}{%
\begin{tabular}{lcccccc}
\toprule
\textbf{Data} & \textbf{Attraction} & \textbf{Hotel} & \textbf{Restaurant} & \textbf{Taxi} & \textbf{Train} & \textbf{Total}\\ 
\midrule
$1\%$  & $106$ & $111$ & $116$ & $105$ & $111$ & $549$ \\
$5\%$  & $547$ & $553$ & $553$ & $548$ & $547$ & $2748$ \\
$10\%$  & $1093$ & $1112$ & $1109$ & $1086$ & $1095$ & $5495$ \\
$\mathrm{synthetic_1}$ & $526$ & $2968$ & $1843$ & $795$ & $1536$ & $7668$\\
$\mathrm{synthetic_5}$  & $2142$ & $9223$ & $6146$ & $2856$ & $5422$ & $25789$ \\ 
\bottomrule
\end{tabular}%
}
\end{center}
\caption{Synthetic data distribution across domain.}
\label{tab:synthDST_dist}
\end{table}

\subsection{Extended Discussion}
\label{appendix:discussion}

\paragraph{Impact of Test Distribution on the Results.}
Figure \ref{fig:domain_dist} depicts the coarse and fine-grained distribution of the different domains in the MultiWOZ test set. The coarse-grained distribution suggests a relatively balanced representation of all domains, except for the \textit{taxi} domain, which is less prominent. However, when examining the fine-grained distribution, a different picture emerges. Since MultiWOZ comprises multiple domains within a single dialogue, some domains overlap. In this fine-grained analysis, it becomes evident that the \textit{attraction} domain, when considered in isolation, is the most underrepresented sub-category. However, it frequently appears in tandem with other domains such as \textit{train} and \textit{restaurant}. Therefore, we hypothesize that an increase in the performance of \textit{train} and \textit{restaurant} results in a decrease in \textit{attraction}. This hypothesis is substantiated by the results presented in Table \ref{tab:results_main}. Specifically, the scores for the \textit{attraction} domain demonstrate an increase, while the\textit{train} and \textit{restaurant}  domains experience a decrease in performance (as evidenced by the Few shot$_\mathrm{random}$). Similarly, the opposite is observed for Few Shot$_\mathrm{synthetic}$.

\paragraph{Impact of Off-The-Shelf Retriever.}
Unlike other ICL approaches, we refrain from fine-tuning the retriever to mimic a no-training data scenario. As illustrated by the results in Table \ref{tab:results_main}, the performance demonstrates little correlation with the expansion of the retrieval pool. Furthermore, there are instances where the performance actually decreases, notably in the $1\%\rightarrow5\%$ setup for synthetic data and the $5\%\rightarrow10\%$ setup for training data across both datasets. We postulate that this might be attributed to off-the-shelf retrievers occasionally retrieving irrelevant examples since they lack awareness of the semantics of the end-task data. In summary, our results attest that we can achieve good performance with a small data set with off-the-shelf retrievers.

\subsection{Human Evaluation}
\label{appendix:human_eval}

\begin{table}[h!]
\begin{center}
\resizebox{\columnwidth}{!}{%
\begin{tabular}{ll}
\toprule
\textbf{Metric} & \textbf{Scale} \\
\midrule
\multirow{5}{*}{Grammar} & 1 = Highly Incoherent or Unintelligible \\
& 2 = Poorly Constructed and Difficult to Understand \\
& 3 = Moderately Fluent, but Some Awkwardness \\
& 4 = Mostly Fluent and Easily Understandable \\
& 5 = Extremely Fluent and Natural \\
\cmidrule(lr){1-2}
\multirow{5}{*}{Coherence} & 1 = Responses Lack Logical Flow and Are Highly Disjointed \\
& 2 = Poor Logical Flow, and Responses Often Do Not Connect \\
& 3 = Responses Have Some Logical Flow but Lack Consistency \\
& 4 = Logical Flow Is Mostly Maintained with Few Disruptions \\
& 5 = Highly Coherent and Smooth Logical Flow \\
\cmidrule(lr){1-2}
\multirow{5}{*}{Naturalness} & 1 = Very Robotic and Unnatural, Clearly Generated \\
& 2 = Lack of Natural Language Patterns, Not Believable \\
& 3 = Moderately Natural, but Still Exhibits Robot-Like Phrasing \\
& 4 = Fairly Natural and Believable in a Conversational Context \\
& 5 = Extremely Natural and Difficult to Distinguish from Human Speech \\
\cmidrule(lr){1-2}
\multirow{5}{*}{Annotations} & 1 = Completely Incorrect \\
& 2 = Partially correct, covering most of the slot value pairs \\
& 3 = Exactly correct, covering all the possible slot value pairs \\
\bottomrule
\end{tabular}%
}
\end{center}
\caption{Human Evaluation Scale.}
\label{tab:human_eval}
\end{table}

\end{document}